\documentclass[11pt,a4paper]{article}
\usepackage{authblk}
\usepackage[hyperref]{acl2020}
\usepackage{times}  
\usepackage{latexsym}
\usepackage{graphics}
\usepackage{adjustbox}

\usepackage{framed}
\graphicspath{{./}}
\usepackage{amsfonts} 
\usepackage{amsmath}
\usepackage[
singlelinecheck=false 
]{caption}

\usepackage{microtype}

\aclfinalcopy 

\title{Multitask Finetuning for Improving Neural Machine Translation in Indian Languages}

\author[1]{\textbf{Shaily Desai}}
\author[1]{\textbf{Atharva Kshirsagar}}
\author[2]{\textbf{Manisha Marathe}}

\affil[1,2]{Department of Computer Engineering, PVG's COET, Affiliated to Savitribai Phule Pune University, India.}

\affil[1]{\texttt {\{shaily.desai21, atharvakshirsagar145\} @gmail.com}}
\affil[2]{\texttt {mvm\_comp@pvgcoet.ac.in }}

\date{} 

\begin{document}
\maketitle 

\date{}
\begin{abstract}
Transformer based language models have led to impressive results across all domains in Natural Language Processing. Pretraining these models on language modeling tasks and finetuning them on downstream tasks such as Text Classification, Question Answering and Neural Machine Translation has consistently shown exemplary results. In this work, we propose a Multitask Finetuning methodology which combines the Bilingual Machine Translation task with an auxiliary Causal Language Modeling task to improve performance on the former task on Indian Languages. We conduct an empirical study on three language pairs, Marathi-Hindi, Marathi-English and Hindi-English, where we compare the multitask finetuning approach to the standard finetuning approach, for which we use the mBART50 model. Our study indicates that the multitask finetuning method could be a better technique than standard finetuning, and could improve Bilingual Machine Translation across language pairs. 
\end{abstract}

\section{Introduction}\label{introduction}
After being introduced in \citep{sutskever2014sequence, cho2014properties, bahdanau2016neural}, Neural Machine Translation(NMT) rapidly replaced and outperformed the traditional Statistical models for translation tasks. It has since achieved state-of-the-art performances for a multitude of languages. This can be attributed to the fact that NMT uses continuous representations for languages, is capable of handling long-distance dependencies, and requires significantly less feature-engineering \citep{TAN20205}. Nearly all such models consist of an encoder-decoder architecture. Earlier applications of this framework incorporate Recurrent Neural Networks \citep{cho-etal-2014-learning} and Convolutional Neural Networks \citep{kalchbrenner2017neural, gehring2017convolutional} as their encoder and decoder components. While there have been many variants of recurrent networks which have performed well in language modeling, they bear a few shortcomings. Most importantly, they inhibit parallelization, and have no explicit model hierarchy. Moreover, training deep neural networks with recurrence is challenging and can result in vanishing or exploding gradients \citep{pascanu2013difficulty}.

The concept of attention was introduced in \citep{bahdanau2016neural} to avoid having a fixed-length source sentence representation, which solved the fixed-length bottleneck problem. The attention mechanism has also eased optimization difficulty, and is considered to be a milestone in Machine Translation research. 

\citealt{vaswani2017attention} introduced the Transformer architecture, which forgo RNNs and CNNs, and are entirely based on the attention mechanism. Transformers proved to be efficacious on sequence-to-sequence tasks, and, as a result, transformer based language models emerged. Pretraining deep transformers on language modeling tasks has significantly improved performances on NLP tasks as compared to training from scratch \citep{devlin2019bert,radford2019language,lewis2019bart,liu2019roberta}. The idea behind pretraining is that the models are initialized with general linguistic knowledge, which can then be applied to downstream tasks by further finetuning the model.

Multitask Learning(MTL) \citep{Caruana2004MultitaskL} has been successful in boosting results across many domains. The increase in performance can be credited to learning shared representations which improves generalization performance in two or more related tasks which have been jointly trained. MTL has also proven advantageous in neural machine translation. \citealt{luong2016multitask} showed that combining Machine Translation with Parsing and Image Captioning led to better translation results. \citealt{niehues2017exploiting} integrated POS tagging, NER and Machine Translation.

In this study, we propose a multitask finetuning methodology which utilises monolingual data to increase the performance of NMT on Indian language pairs. Inspired by the works of \citep{wang2020multitask} and \citep{domhan-hieber-2017-using}, we propose finetuning of a pretrained model on bilingual parallel data and one auxiliary task- Causal Language Modeling (CLM) on the monolingual corpora of the source side language as well as the target side language. We use a pretrained mBART \citep{liu-etal-2020-multilingual-denoising} for multi-task finetuning, and compare the performance with the standard finetuning method on the same model and corpus.

The rest of the paper is organised as follows: Section \ref{data} provides an overview of the dataset used for this study, Section \ref{method} describes the proposed methodology in depth. In Section \ref{experiment}, we describe our experimental setup and the results are discussed in Section \ref{results}. Section \ref{conclusion} provides a conclusion to the study and discusses possible future works.

\section{Dataset} \label{data}
For translation, we selected three language pairs: Marathi-English,  Hindi-English and Marathi-Hindi. We chose these languages as Hindi, English and Marathi are three of the four most spoken languages across India. The parallel corpus that we used was \emph{Samanantar} \citep{ramesh2021samanantar} which is the largest publicly available parallel corpora collection for Indic languages. The Samanantar corpus has 1.99 million sentence pairs for Marathi-Hindi, 3.32 million sentence pairs for Marathi-English, and 8.56 million sentence pairs for Hindi-English. We randomly selected a subset of these examples for translation. The distribution of this subset is given in Table \ref{table1}\footnote{xx in Table \ref{table1} signifies two target languages in every case. For example, in Mr\textrightarrow xx, xx is English and Hindi.}.
\par The data used for the language modeling task was sampled from \emph{IndicCorp} \citep{kakwani-etal-2020-indicnlpsuite}, which is a monolingual corpora spanning 11 Indic languages. Out of these we select just our source and target side languages i.e Hindi, Marathi and English. We sample a subset of IndicCorp for each of these languages. The distribution for monolingual data selected for the Causal Language Modeling is given in Table \ref{table2}.

\begin{table}[htp]
\centering
\resizebox{180pt}{37pt}{
\resizebox{\textwidth}{!}{%
\begin{tabular}{cccccc}
\hline\hline  & \textbf{Mr\textrightarrow xx} &\textbf{Hi\textrightarrow xx}
&\textbf{En\textrightarrow xx}\\
\hline\hline
\textbf{Training} & 100k & 100k & 100k\\
\textbf{Validation} & 20k & 20k & 20k\\
\textbf{Testing} & 5k & 5k & 5k\\
\hline\hline
\end{tabular}}
}
\caption{\label{Bilingual-corpus}Data distribution of bilingual parallel corpora}\label{table1}
\end{table}

\begin{table}[h]
\centering
\resizebox{165pt}{40pt}{
\begin{tabular}{ccc}
\hline\hline & \textbf{Total Available} & \textbf{Selected}\\
 & \textbf{Corpora} &\\
\hline\hline
\textbf{Mr} & 34.0M & 70k\\
\textbf{Hi} & 63.1M & 70k\\
\textbf{En} & 54.3M & 70k\\
\hline\hline
\end{tabular}
}
\caption{\label{Monolingual-corpus}Data distribution of monolingual corpora}\label{table2}
\end{table}

\section{Methodology} \label{method}
We use the pretrained mBART50 model in a multitask setting, with translation as the main task and self-supervised language modeling as an auxiliary task. We then compare the performance of this model to that of the conventionally finetuned mBART50 which has been trained solely on the translation task. The principal components of the multitask model are briefly explained in this section.
\subsection*{Multitask Learning}
Translation is the primary downstream task in our multitask model, for which we train a bitext corpus $\mathnormal{D_B}$ which consist of sentence pairs $\mathnormal(s,t)$, and is optimized on the crossentropy loss function:

\begin{equation}
\mathcal{L_\mathnormal{T}} = 
\mathbb{E_{\mathnormal{(s,t)\sim\mathnormal{D_B}}}} [-\log{P(t|s)}]
\end{equation}

Where $\mathnormal(s,t)$ represents the source and target text respectively. As a large amount of monolingual corpora for these languages is available, we leverage it to improve NMT performance by training language modeling auxiliary tasks with our primary translation task. We train the source as well as target side languages with their respective monolingual corpora on the Causal Language Modeling objective.

\subsection*{Causal Language Modeling}
In Causal Language Modeling(CLM), the model has to predict the next token given a sequence of previous tokens. Given a monolingual corpus $\mathnormal{D_M}$, CLM minimizes the crossentropy loss:

\begin{equation}
\mathcal{L_\mathnormal{CLM}} = 
\mathbb{E_{\mathnormal{(x)\sim\mathnormal{D_M}}}} [-\log{P(x_t|x_{t-1},x_{t-2},\cdot\cdot\cdot, x_1)}]
\end{equation}

\normalsize
Where $\mathnormal{x_t}$ is the token predicted given $\mathnormal{(x_{t-1},\cdot\cdot x_1)}$ tokens. CLM has proven to be highly effective in enhancing sequence generation and natural language understanding \citep{radford2019language}. We thus explore it’s efficacy and leverage it as an auxiliary task in our multitask framework. 
\subsection*{Training}
Both the Translation and Causal Language Modeling objective are trained jointly and the cross entropy losses for both tasks are added together:
\begin{equation}
\mathcal{L_\mathnormal{MTL}} = \mathcal{L_\mathnormal{T}} + \mathcal{L_\mathnormal{CLM}}
\end{equation}

\section{Experimental Setup}\label{experiment}
We use a standalone mBART50 model as the baseline to compare results on Machine Translation with the multitask methodology proposed in section \ref{method}. We chose an mBART50 model which uses a standard sequence-to-sequence Transformer architecture with 12 encoder and decoder layers each. This model has been shown to perform relatively well on machine translation tasks in multiple Indian languages.
\subsection*{Baseline Models(mBART50)}
We make use of the pretrained Huggingface {\fontfamily{qcr}\selectfont Transformers
}\footnote{\url{https://huggingface.co/transformers/}} library implementation for mBART50-large. This model was finetuned on the parallel corpus described in Section \ref{data}. The batch size chosen for our baseline models is $16$. We finetune a separate pretrained mBART50 for each language pair in our parallel dataset (for example, En\textrightarrow Mr and Mr\textrightarrow En constitute different models).

\subsection*{Multitask Models(MTL-mBART50)}
We use the same mBART50 implementation as that of the baseline for our multitask models.The {\fontfamily{qcr}\selectfont Transformers
} library does not support multitask learning; Hence, we employ an approach inspired by a publicly available implementation for the same\footnote{\url{https://github.com/zphang/zphang.github.io/blob/master/files/notebooks/Multi_task_Training_with_Transformers_NLP.ipynb}}. For the sake of a direct comparison between the Neural Machine Translation performance between Baseline models and the multitask models, we use the same parallel corpus as that of the baselines. Additionally, the multitask models are trained for the auxiliary task of Causal Language modeling on the monolingual data described in Section \ref{data}. Due to computational constraints, the batch size selected for the multitask models was $2$. Similar to the baseline models, a separate multitask model was trained for each language pair considered.

\begin{figure}[h!]
  \includegraphics[width=0.5\textwidth,height=10cm]{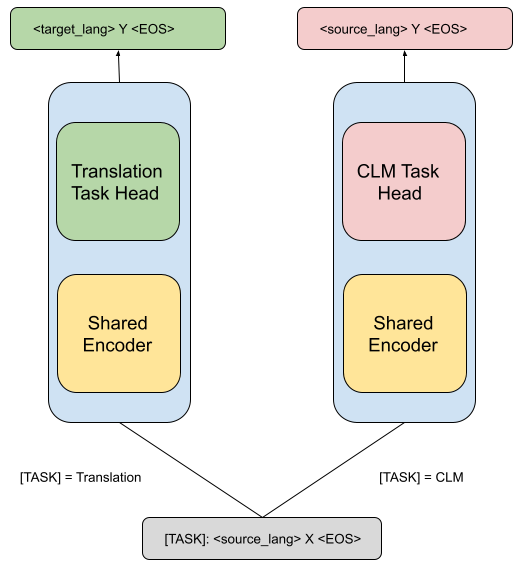}
  \caption{A multitask setup using the mBART model from the Huggingface {\fontfamily{qcr}\selectfont Transformers
} Library. The two models above share an encoder, and have different decoders and therefore, make one multitask model. Here, $X$ = input tokens and $Y$ = generated output tokens.}
\end{figure}

In both the baseline and multitask models, we freeze the first $6$ of mBART50’s $12$ encoder layers. All models were trained for one epoch each on one P100 GPU provided by Google Colab \footnote{\url{https://colab.research.google.com}}. The models were trained using the Adam optimizer \citep{kingma2017adam} with $\beta_1 = 0.9$ and $\beta_2 = 0.999$. The learning rate was kept constant at $1e-5$ across the training run.

During inference, we decoded the generated sentences with a beam of size $2$ and used a length penalty of $1.2$. We then measure and report the BLEU \citep{papineni-etal-2002-bleu} scores calculated after applying the Smoothing Function method 4 \footnote{\url{https://www.nltk.org/_modules/nltk/translate/bleu_score.html}} in the {\fontfamily{qcr}\selectfont nltk
} \citep{bird-loper-2004-nltk} library.

\begin{table*}[!htb]
\captionsetup{justification=centering,margin=1cm}
\centering
\resizebox{0.85\textwidth}{30pt}{
\begin{tabular}{lllllllll}
\hline \textbf{Model} & &\textbf{Mr\textrightarrow Hi} & \textbf{Hi\textrightarrow Mr} & \textbf{Mr\textrightarrow En} & \textbf{En\textrightarrow Mr} &
\textbf{En\textrightarrow Hi} & 
\textbf{Hi\textrightarrow En}\\ \hline\hline
\textbf{Baseline-mBART50}& & 9.48 & 5.61 & 10.17 & 5.33 & 6.49 & 8.12\\
\hline\textbf{MTL-mBART50}& & \textbf{10.33} & \textbf{6.85} & \textbf{11.84} & \textbf{6.47} & \textbf{8.71} & \textbf{9.17}\\
\hline\hline
\end{tabular}
}
\caption{\label{performance-comparison}Resulting BLEU scores on different language pairs for the baseline models and our models.}\label{table3}
\end{table*}

\section{Results}\label{results}
Table \ref{table3} shows the cumulative 4-gram BLEU scores of the baseline as well as the multitask finetuned models on different language pairs. The multitask methodology, when trained on the same parallel corpus as the baseline models, experiences a $10-20\%$ improvement in BLEU scores, and this increase in the metric is consistently seen across evaluations for all the language pairs considered.

We ascribe this improvement in BLEU scores to our multitask models' ability to better generate sentences and their increased Natural Language Understanding which is a result of training the Machine Translation task in conjunction with CLM for both, source and target side languages. It would be reasonable to postulate that joint training for the two aforementioned tasks facilitated the generation of more coherent translations, which, in comparison to the baseline model translations, were more similar to the ground truths in the parallel corpus, resulting in better BLEU scores.

Considering the amount of bitext data on which both the model variants were trained, the resulting BLEU scores are expectedly low. But since the purpose of this study is to estimate the viability of our proposed method, the increment seen in scores from the multitask models adequately confirm our hypothesis.

\section{Conclusion}\label{conclusion}
In this work we propose a Multi-Task finetuning methodology for Bilingual Neural Machine translation, which, along with training a model on a bilingual parallel corpus, also trains it on a Causal Language Modeling objective for both the source and target side monolingual data in a self supervised manner. We show that this approach outperforms the standard Fine-tuning methodology for Neural Machine Translation for the considered language pairs. 

This study is of preliminary nature and for future work we aim to train other transformer based language models using the same methodology. Due to the modest computational resources available to us, we were compelled to train on relatively low amounts of data. Therefore, we also hope to use this method to train on larger bitext and monolingual corpora, for more Indian Language pairs. Although the proposed methodology has proven to be effective in our experiments, training models on larger datasets using this multitask framework would add to its credibility.

\bibliography{references}
\bibliographystyle{acl_natbib}
\end{document}